\documentclass[letterpaper]{article} 
\usepackage[]{aaai25}  
\usepackage{times}  
\usepackage{helvet}  
\usepackage{courier}  
\usepackage[hyphens]{url}  
\usepackage{graphicx} 
\usepackage{natbib}  
\usepackage{caption} 
\usepackage{algorithm}
\usepackage{algorithmic}
\usepackage{amsmath}
\usepackage{amssymb}
\usepackage{tabularray}
\usepackage{verbatim}
 \usepackage{multirow}
 \usepackage{vcell}
\usepackage{newfloat}
\usepackage{listings}
\DeclareCaptionStyle{ruled}{labelfont=normalfont,labelsep=colon,strut=off} 
\floatstyle{ruled}
\newfloat{listing}{tb}{lst}{}
\floatname{listing}{Listing}
\title{Zero-Shot NAS via the Suppression of Local Entropy Decrease}
\author{
    Ning Wu\textsuperscript{\rm 1},
    Han Huang\textsuperscript{\rm 1},
    Yueting Xu\textsuperscript{\rm 1},
    Zhifeng Hao\textsuperscript{\rm 2}
}
\affiliations{
    \textsuperscript{\rm 1}School of Software, South China University of Technology\\
    \textsuperscript{\rm 2}Department of Mathematics, College of Science, Shantou University\\
    wuningscut@163.com, hhan@scut.edu.cn,  yuetingxu96@163.com, haozhifeng@stu.edu.cn

}

\usepackage{bibentry}
\newtheorem{example}{Example}

\newtheorem{definition}{Definition}

\newtheorem{proposition}{Proposition}

\begin{document}

\maketitle

\begin{abstract}
Architecture performance evaluation is the most time-consuming part of neural architecture search (NAS). Zero-Shot NAS accelerates the evaluation by utilizing zero-cost proxies instead of training. Though effective, existing zero-cost proxies require invoking backpropagations or running networks on input data, making it difficult to further accelerate the computation of proxies. To alleviate this issue, architecture topologies are used to evaluate the performance of networks in this study. We prove that particular architectural topologies decrease the local entropy of feature maps, which degrades specific features to a bias, thereby reducing network performance. Based on this proof, architectural topologies are utilized to quantify the suppression of local entropy decrease (SED) as a data-free and running-free proxy. Experimental results show that SED outperforms most state-of-the-art proxies in terms of architecture selection on five benchmarks, with computation time reduced by three orders of magnitude. We further compare the SED-based NAS with state-of-the-art proxies. SED-based NAS selects the architecture with higher accuracy and fewer parameters in only one second. The theoretical analyses of local entropy and experimental results demonstrate that the suppression of local entropy decrease facilitates selecting optimal architectures in Zero-Shot NAS.
\end{abstract}

\section{Introduction}
\label{Introduction}
Neural architecture search (NAS) algorithms automate the process of designing architectures, overcoming the limitations in terms of manual design efficiency. Nevertheless, the evaluation of networks relies on computationally expensive network training, which makes NAS dependent on high-performance GPUs \cite{RN466, RN467, RN427}. Numerousds studies have been conducted to improve the search speed of NAS metho \cite{RN459, RN460, RN461, RN462, RN487}.	Zero-Shot NAS \cite{RN428} is a promising paradigm in accelerating network performance evaluation. It leverages geometric features derived from the network parameters or gradient landscape as evaluation criteria without complete training of networks.

Challenges persist despite the effectiveness of existing zero-cost proxies.   
(i) Current zero-cost proxies run networks or invoke backpropagations, which is time-consuming for large search spaces (e.g., as shown in Example~\ref{exam:1}). (ii) Most proxies rely on the input data. The reliance on input data inadvertently underestimates the significance of architecture properties. To overcome these challenges, we propose a topology-based proxy to extract the architecture properties as evaluation criteria.

\textbf{Objectives.}   The overriding objective of this study is to develop a zero-cost proxy distinguished by its speed and reduced consumption of floating-point operations. This proxy is supposed to be data-free and network-running-free.

\begin{figure*}[!htp]
 	\centering
 	\includegraphics[width=1\textwidth]{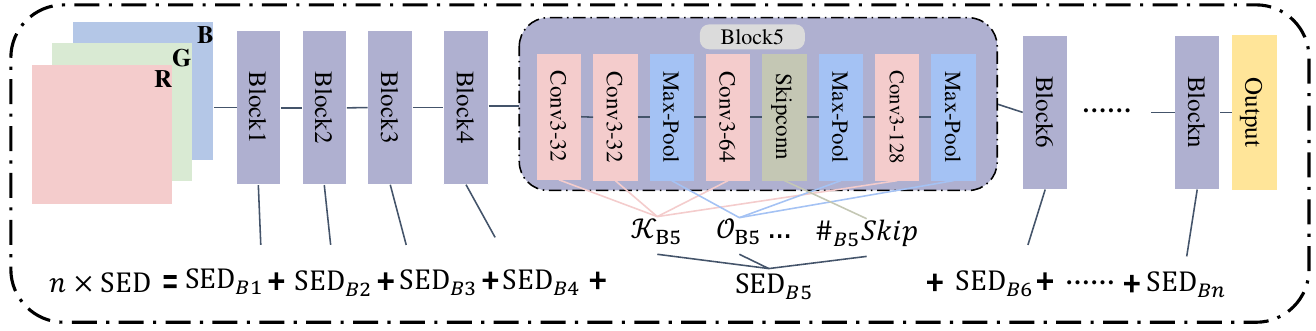}
 	\caption{Schematic diagram of calculating SED.}
 	\label{fig:arc}
 \end{figure*}

\textbf{Method and Results.}  From the perspective of entropy decrease, this study explores the impact of network architecture on the local entropy of feature maps. We prove that decreasing the local entropy of feature maps degrades specific features to biases, reducing network performance. Furthermore,  irrational operation settings (including convolution, pooling, and skip connection) are proven to trigger local entropy decrease. Based on the above analyses,  the suppression of local entropy decrease  (SED) proxy is proposed to quantify the suppression of local entropy decrease as a proxy for network performance. The computation of SED, which is data-free and network-running-free, relies solely on architecture topologies to accelerate the evaluation. A schematic of SED calculation is provided in Figure~\ref{fig:arc}.  Experimental results demonstrate that SED outperforms most state-of-the-art (SOTA) zero-cost proxies on multiple benchmarks. In terms of efficiency,  SED significantly accelerates architecture evaluation, taking only \textbf{3.4e-5} seconds to evaluate an architecture in NATS-Bench-TSS. In contrast, existing SOTA methods (e.g., grad\_norm \cite{RN432}) take at least 0.54 seconds. In terms of architecture selection,  SED selects architectures ranked \textbf{3rd}, \textbf{1st}, and \textbf{24th} for NATS-Bench-TSS using CIFAR-10, CIFAR-100, and ImageNet16-120, respectively.
The Spearman's $\rho$ of SED achieves 0.09 higher than the SOTA proxy on NAS-Bench-301 out of 2,000 randomized architectures. Moreover, the SED-based NAS selects the architectures with the highest accuracy among the compared proxies in complete NAS tasks. For example, SED-based NAS selects architecture with 81.09\% test accuracy, which is 0.42\% higher than the SOTA proxies.

\textbf{Contributions.} Compared with previous work, this study utilizes network architecture topologies to construct the proxy rather than network parameters and input data. The main contributions of this study are as follows:
\begin{itemize}
	\item Theoretically, we prove that specific network architectures reduce local entropy. Moreover, for classification tasks, entries in feature maps degrade to biases as the local entropy decreases,  thereby reducing network performance. 
	
	\item Based on the theoretical analyses, the suppression of local entropy decrease caused by architectures is quantified as a proxy for network performance. This proxy outperforms SOTA proxies on multiple benchmarks, reducing the time consumed in architecture evaluation by three orders of magnitude compared to SOTA proxies.
	
\end{itemize}

\begin{example}
	\label{exam:1}
	Evaluation of a single architecture in the NATS-Bench-TSS benchmark using grad\_norm \cite{RN432} on a 3090 GPU takes only 0.54 seconds. NATS-Bench-TSS consists of 4 nodes and 5 related operations, totaling  15,625 architectures, and it takes 2.3 GPU hours to evaluate all the architectures in NATS-Bench-TSS. The architecture space with 5 nodes and 5 related operations, using the exact construction mechanism as NAS-Bench-TSS, contains 9,765K architectures. However, evaluating this architecture space takes at least 1,480 GPU hours based on grad\_norm.
\end{example}

\section{Related Work}
In this section, we briefly introduce the existing research on zero-cost proxies.
Gradient descent is a fundamental method used to update neural network parameters. Therefore, zero-shot proxies utilize gradients to evaluate network performance. A straightforward method is to use the $p$-norm of parameter gradients as a proxy \cite{RN432}.
The snip proxy \cite{RN433} considers gradient backpropagation and forward inference information. Based on snip, synflow \cite{RN434} considers the sign of the inner product as an improvement.
First-order derivatives only measure the magnitude of the change in parameters. The grasp \cite{RN435} proxy uses the second-order derivatives to capture the rate of parameter change. Moreover, synflow and GraSP are proved to be the different approximations of neural networks' first-order Taylor expansion \cite{RN435, RN436}. Moreover, the fisher \cite{RN438, RN439} proxy was proposed as an approximation of the second-order Taylor expansion of the networks. GradSign proxy \cite{RN437} works as an approximation of training loss to predict the network performance. In addition to the use of gradient over parameters, Jacobcov proxy \cite{RN454, RN455} uses the gradient over the input data. Zico \cite{RN420} adopts this idea and analyzes the effect on the convergence of neural networks caused by different samples. The NTK condition number \cite{RN440, RN453} utilizes the eigenvalues of the samples' Jacobian matrix.     

Gradient calculation and backpropagation are time-consuming, which makes gradient-based proxies computationally expensive. Gradient-free proxies have been proposed for faster evaluation.
The number of linear regions of a neural network represents the total number of parts that the network can divide its input space. This is directly related to the Vapnik-Chervonenkis dimension \cite{RN441}. Therefore, the number of linear regions represents the generalizability of a particular network \cite{RN442, RN443, RN444}. Considering the relative position between different linear regions, s-score \cite{RN455} introduces the Hamming distance as an improvement. The s-score is further integrated with the NASWOT algorithm. The network architecture topologies are also used to build proxies, including $\beta$-score \cite{RN576}, NN-Mass \cite{RN445}  and NN-Degree \cite{RN446}. Zen-score \cite{ming_zennas_iccv2021} was proposed to rank architectures in a few forward inferences. The Pearson correlation matrix of the feature map is utilized by meco \cite{RN411}  to eliminate the dependence on gradients.  
Expressivity and trainability are considered in ETAS \cite{RN580}, while AZ-nas \cite{RN578} adds the consideration of progressivity and complexity.
The SWAP score \cite{pengswap}, provide a novel insight form statistical analysis.

\section{SED: Suppression of Local Entropy Decrease }

\subsection{Preliminaries}
\textbf{Notations.}  \textbf{(We will improve the mathematical expression in the next version, for more clear presentation.)} The architecture space is denoted as $\mathcal{A}$. Each architecture $A\in \mathcal{A}$  consists of a sequence of blocks $(B_1,B_2,\cdots,B_n)$. 
 $\mathcal{OPT}(A)$ consists of all the optional operations in $A$.  
A convolutional kernel  is a tensor denoted as $\mathbf{K}\in \mathbb{R}^{k_w \times k_h \times k_{c}}$. The pooling operation is denoted by $\mathbf{O}$ (e.g., $\mathbf{O}_{max}$ is the maximum pooling operation). 
The size of $\mathbf{O}$  is denoted by the number of features involved in a polling operation.  The vector $\mathbf{s}\in \mathbb{N}^{3}$ is used to represent the stride.   The input of the $l$-th layer of the network is denoted by  $\mathbf{X}^{(l)}$.  
$[\mathbf{X}]_{i,j,v}$ is the $(i,j,v)$-th entry of $\mathbf{X}$.   
We use the notation $'*'$ as a discrete convolution operator. The convolution of tensors $\mathbf{X}\in\mathbb{R}^{k_w\times k_h\times k_c}$ and $\mathbf{K}$ is defined as $(\mathbf{X}* \mathbf{K})(k_w,k_h,k_c)=\sum\limits_{i}\sum\limits_{j}\sum\limits_{v}[\mathbf{\mathbf{X}}]_{i,j,v}\mathbf{K}_{k_w+1-i,k_h+1-j,k_c+1-v}$.  
$\mathbb{S}_{\mathbf{X}}$ is a set that consists of all entries in $\mathbf{X}$.   An multivariate Gaussian distribution is represented by $\mathcal{N}_{n}(\boldsymbol{\mu},\boldsymbol{\Sigma})$, where $\boldsymbol{\mu}$ and $\boldsymbol{\Sigma}$ denote the expected value and covariance matrix of $n$ variables, respectively.  

The entropy discussed in this study includes one-dimensional entropy and multivariate Gaussian entropy, as defined below. 
\begin{definition}[One-dimensional Entropy] 
	\label{O_E}
	Given a tensor $\mathbf{X}$, its one-dimensional entropy is defined as the following equation: \[\mathcal{H}_{1}(\mathbf{X})=-\sum\limits_{x\in \mathbb{S}_{\mathbf{X}}}\mathrm{P}_{\mathbf{X}}(x)\log_2\mathrm{P}_{\mathbf{X}}(x)\] 
	where $\mathrm{P}_{\mathbf{X}}(\cdot)$ calculates the frequency at which an element occurs in $\mathbf{X}$. 
\end{definition}

\begin{definition}[Entropy of Multivariate Gaussian] 
	\label{M_E}
	Given a tensor of variables $\mathbf{X}\sim \mathcal{N}_{whc}(\boldsymbol{\mu},\boldsymbol{\Sigma})$, the entropy of $\mathbf{X}$ is defined as:
	\[\mathcal{H}_{2}(\mathbf{X})=-\int \underset{}{\mathrm{P}(\mathbf{X})}\log_2\underset{}{\mathrm{P}(\mathbf{X})} \mathrm{d}\mathbf{X}\]
	 where $\mathrm{P}(\cdot)$ is the probability density function  of the multivariate Gaussian distribution. 
\end{definition}	
Convolution is a process of extracting information by reducing the entropy. Moreover, the convolution operation is performed on a portion of the input at a time.   For ease of representation, this input partition is defined as a subtensor with the help of set $\mathbb{Q}$.  $\mathbb{Q}_{m,n}$ denotes the set of all sequences of $m$ integers where $m$ and $n$ are integers with $1\leq m \leq n$. The elements in $\mathbb{Q}_{m,n}$ are strictly increasing and chosen from $\{1,2,3,\cdots,n\}$.

\begin{definition}[Subtensor] Suppose a tensor  $\mathbf{X}\in \mathbb{R}^{w\times h \times c}$. $m,n$ and $r$ are positive integers and $1\leq m \leq w$, $1\leq n \leq h$, $1\leq r \leq c$. $\alpha \in \mathbb{Q}_{m,w}$, $\beta \in \mathbb{Q}_{n,h}$, $\gamma\in\mathbb{ }Q_{r,c}$.  We assign $\alpha=(i_1,i_2,\cdots,i_m)$, $\beta=(j_1,j_2,\cdots,j_n)$, $\gamma=(v_1,v_2,\cdots,v_r)$. The subtensor$\mathbf{X}[\alpha;\beta;\gamma]$  of $\mathbf{X}$  is a tensor of size $m\times n\times r$, where $[\mathbf{X}[\alpha;\beta;\gamma]]_{a,b,e}=[\mathbf{X}]_{i_a,j_b,v_e}$ for $1\leq a\leq m$, $1\leq b\leq n$, $1\leq e\leq r$.
\end{definition}

For the convolution operation between a convolutional kernel $\mathbf{K}$ and a subtensor $\mathbf{X}^{(l)}[\alpha_{i,k_w};\beta_{j,k_h};\gamma_{v,k_c}]$ of size $k_w\times\ k_h\times k_c$, 
we define that $\alpha_{i,k_w}=(1+(i-1)s_1-\lfloor\frac{k_w}{2}\rfloor,\cdots,1+(i-1)s_1,\cdots,1+(i-1)s_1+\lfloor\frac{k_w}{2}\rfloor)$, $\beta_{j,k_h}=(1+(j-1)s_2-\lfloor\frac{k_h}{2}\rfloor,\cdots,1+(j-1)s_2,\cdots,1+(j-1)s_2+\lfloor\frac{k_h}{2}\rfloor)$, and $\gamma_{v,k_c}=(1+(v-1)s_3-\lfloor\frac{k_c}{2}\rfloor,\cdots,1+(v-1)s_3,\cdots,1+(v-1)s_3+\lfloor\frac{k_c}{2}\rfloor)$.
We abbreviate $\mathbf{X}^{(l)}[\alpha_{i,k_w};\beta_{j,k_h};\gamma_{v,k_c}]$ to $\mathbf{X}^{(l)}[\alpha_{i,k_w};\beta_{j,k_h}]$ when the convolution kernel and the input data have the same number of channels.

\subsection{The Construction of SED}
In this subsection,  the local entropy of feature maps was used to connect network architecture topologies and network performance to construct the SED proxy.

\subsubsection{Relation between Local Entropy and Network Performance}
\label{Effect of Local Entropy on CNN Performance}
We analyze the mechanism by which local entropy causes the decrease in network performance and formalize this mechanism in Proposition~\ref{pro1} and Proposition~\ref{pro1_multiple}.

\begin{proposition}
	\label{pro1}
	For the classification task, given a feature map $\mathbf{X} \in \mathbb{R}^{w\times h\times 1}$, a convolution kernel $\mathbf{K}\in \mathbb{R}^{k_w\times k_h\times 1}$ and stride  $\mathbf{s}=(s_1,s_2,0)$, where $k_w$ and $k_h$ are odd and less than $w$ and $h$, respectively. The convolution operation $ (\mathbf{X}[\alpha_{i,k_w};\beta_{j,k_h}]*\mathbf{K})(k_w,k_h,k_c)$ works as a bias of $\mathbf{K}$, when $\mathcal{H}_{1}(\mathbf{X}[\alpha_{i,k_w};\beta_{j,k_h}])=0$  under the rectified linear unit activation function.
\end{proposition}

Proposition~\ref{pro1} suggests that subtensor $\mathbf{X}[\alpha_{i,k_w};\beta_{j,k_h}]$, with zero local one-dimensional entropy, provides inefficient feedback for convolution kernels training.
We extend this result to the entropy of multivariate Gaussian in Proposition~\ref{pro1_multiple}.

\begin{proposition}
	\label{pro1_multiple}
	For the classification task, given a feature map $\mathbf{X}\sim \mathcal{N}_{wh}(\boldsymbol{\mu}_2,\boldsymbol{\Sigma}_2)$, a convolution kernel $\mathbf{K}\in \mathbb{R}^{k_w\times k_h\times 1}$ and stride  $\mathbf{s}=(s_1,s_2,0)$, where $k_w$ and $k_h$ are odd and less than $w$ and $h$, respectively.  The convolution operation $(\mathbf{X}[\alpha_{i,k_w};\beta_{j,k_h}*\mathbf{K})(k_w,k_h,k_c)$ works as a constant, when $\mathcal{H}_{2}(\mathbf{X}[\alpha_{i,k_w};\beta_{j,k_h}])\to -\infty$  under the rectified linear unit activation function.
\end{proposition}

Proposition~\ref{pro1_multiple} reveals that increasing the multivariate Gaussian entropy of the subtensor of feature maps facilitates the improvement in network performance. The analyses in this subsection demonstrate that small local entropy reduces network performance. The detailed proofs are presented in the Appendix (in Proof~2 and  Proof~3 ).

\subsubsection{Relation between Local Entropy and Network Architecture}
\label{Effect of Architecture Topology  on  Local Entropy} 
In this subsection, we analyze the mechanism by which network operations prevent a decrease in the local entropy of feature maps. The mechanism is presented as  Proposition~\ref{pro2}, Proposition~\ref{pro3}, and Proposition~\ref{pro4}. 

\begin{proposition}
	\label{pro2}
	For the maximum pooling $\mathbf{O}_{max}$ of size ${o_w\times o_h \times 1}$  taking the feature map $\mathbf{X}^{(l)}\in \mathbb{R}^{ w\times h\times 1}$  as input and $\mathbf{s}=(s_1,s_2,0)$ as stride, where $o_w$ and $o_h$ are less than $w$ and $h$, respectively. At the probability greater than $1-\frac{2(\max\{o_w,o_h\}-2)(w+h)}{wh}$, there exists at least one subtensor $\mathbf{X}^{(l+1)}[\alpha_{i,\lceil o_w/s_1\rceil };\beta_{j,\lceil o_h/s_w\rceil}]$ such that $\mathcal{H}_{1}(\mathbf{X}^{(l+1)}[\alpha_{i,\lceil o_w/s_1\rceil };\beta_{j,\lceil o_h/s_w\rceil}])=0$.
\end{proposition}
Proposition~\ref{pro2} indicates that large pooling kernels will produce subtensors with zero local one-dimensional entropy. Following  Proposition~\ref{pro1}, it degrades specific features in the following convolution operation to a bias when $k_w\leq \lceil o_w/s_1\rceil, k_h\leq \lceil o_h/s_2\rceil$.   A large-size convolution kernel avoids the feature degeneration caused by pooling operators \cite{RN430, RN431}.
From the perspective of local multivariate Gaussian entropy, we explain the phenomenon that larger convolution kernels conduce to better performance. Proposition~\ref{pro3} formalizes the explanation.

\begin{proposition}
	\label{pro3}
	For the given convolution kernel $\mathbf{K}_1 \in \mathbb{R}^{k_{w1}\times k_{h1}\times k_{c1}}$,  $\mathbf{K}_2 \in \mathbb{R}^{k_{w2}\times k_{h2}\times k_{c2}}$, and the feature map $\mathbf{X}\sim \mathcal{N}_{whc}(\boldsymbol{\mu},\boldsymbol{\Sigma})$,  $\mathcal{H}_{2}(\mathbf{X}[\alpha_{i,k_{w1}};\beta_{j,k_{h1}};\gamma_{v,k_{c1}}])\geq \mathcal{H}_{2}(\mathbf{X}[\alpha_{i,k_{w2}};\beta_{j,k_{h2}};\gamma_{v,k_{c2}}] )$ when $k_{w1}k_{h1} k_{c1}\geq k_{w2}k_{h2} k_{c2}$ and $k_{w1}k_{h1} k_{c1}\neq 2k_{w2}k_{h2} k_{c2}$. 
\end{proposition}

Proposition~\ref{pro3} shows that large-size convolution kernels are more favorable for increasing the local multivariate Gaussian entropy. Similarly, skip connections also prevent the decrease of local entropy. Proposition~\ref{pro4}  supports this view.

\begin{proposition}
	\label{pro4}
	For the given  feature map $\mathbf{X}^{(l_1)}\sim \mathcal{N}_{whc}(\boldsymbol{\mu}_1,\boldsymbol{\Sigma}_1)$ and feature map $\mathbf{X}^{(l_2)}\sim \mathcal{N}_{whc}(\boldsymbol{\mu}_2,\boldsymbol{\Sigma}_2)$,
	$\mathcal{H}_{2}((\mathbf{X}^{(l_1)}+\mathbf{X}^{(l_2)})[\alpha_{i,k_w};\beta_{j,k_h}])>\max\{\mathcal{H}_{2}(\mathbf{X}^{(l_1)}[\alpha_{i,k_w};\beta_{j,k_h}]),\mathcal{H}_{2}(\mathbf{X}^{(l_2)}[\alpha_{i,k_w};\beta_{j,k_h}])\}$ holds for any $\alpha_{i,k_w}$ and $\beta_{j,k_h}$.  
\end{proposition}

Skip connections suppress the local entropy decrease, but overfull skip connections lead to redundant information, thereby misleading the optimization direction. The number of channels works in the same  mechanism as the skip connection. Proposition~\ref{pro4} provides an optional explanation for the performance of ResNet \cite{RN457} and GoogleNet \cite{RN458}.
In this section, we prove that irrational network architectures cause a decrease in the local entropy of feature maps, which reduces network performance. 	The proofs of propositions in this section are presented  in the Appendix (in  Proof~3, Proof~4 and Proof~5).

\subsubsection{Topology-based zero-cost proxy: SED}
\label{sec:CSED}   
 Propositions~\ref{pro1} and Propositions~\ref{pro1_multiple} demonstrate that a small local entropy causes a reduction in network performance.  Moreover, Propositions~\ref{pro2}, Propositions~\ref{pro3} and Propositions~\ref{pro4} demonstrate that network architecture topologies affect the value of local entropy according to the following rules:
\begin{itemize}
	\item Architectures with large pooling kernels and small convolution kernels cause a one-dimensional entropy decrease. (According to Proposition~\ref{pro2})
	\item Convolution kernels of large size facilitate obtaining large multivariate Gaussian entropy. (According to Proposition~\ref{pro3})
	\item Skip connections suppress the decrease of multivariate Gaussian entropy. However, excessive skip connections bring redundant information. (According to Proposition~\ref{pro4}) 
\end{itemize}	
\newcommand{\indicator}[1]{\mathbf{I}_{\{#1\}}}
Based on the above rules, we quantify the suppression of local entropy decrease  by network architectures. $\mathcal {K}_{A}$ and $\mathcal {O}_{A}$ represent the set of optional convolution and pooling kernels in $\mathcal {OPT}(A)$, respectively.   $\mathbf{K}>\mathbf{O}$ is used to present $k_w> \lceil o_w/s_1\rceil$ and $k_h> \lceil o_h/s_2\rceil$. We assume that $\mathcal{K}_A$ and $\mathcal{O}_A$ are ordered, i.e., for $\mathbf{K}_i,\mathbf{K}_{i+1}\in \mathcal{K}_A$ and $\mathbf{O}_i, \mathbf{O}_{i+1}\in \mathcal{O}_A$, $\mathbf{K}_i$ and $\mathbf{O}_i$ act in a single operation with the involved features larger than $\mathbf{K}_{i+1}$ and $\mathbf{O}_{i+1}$, respectively. $\#_{A}op$ indicates the number of  $op \in \mathcal{OPT}(A)$, (e.g., $\#_{A}$skip represents the number of skip connections in $A$).   
$\#F_{in}$ and $\#F_{out}$ denote the number of input and output entries in a channel, respectively. The number of input and output channels is represented by $\#C_{in}$ and $\#C_{out}$, respectively. $\indicator{e}=1$ if the event $e$ happen, otherwise $\indicator{e}=0$.   The $sig(\cdot)$ function is defined by $sig(x)\triangleq \frac{1}{1+e^{-x}}, x\in \mathbb{R}$. 

\begin{definition}[SED] Given a network architecture $A\in \mathcal{A}$ consisted of blocks $(B_1,B_2,\cdots,B_n)$, its SED is calculated by the following formula:
	\small
	\begin{align}
	&\	skip\_SED(A)  =    sig(\#_{A}skip)\#_{A}skip  \\
	&\	conv\_SED(A) = \sum\limits_{i=1}^{+\infty}\indicator{\mathbf{K}_{i}\geq \mathbf{O}_{1}} sig(\#_{A}C_{out})\#_{A}\mathbf{K}_i\\
	&	pool\_SED(A) =\left( \sum\limits_{i=1}^{+\infty}\indicator{\mathbf{K}_{i}\geq \mathbf{O}_{1}}\#_{A}\mathbf{K}_i \right)^2+ \left( \sum\limits_{op\in \mathcal {OPT}} \#_{A}op \right)^2 \nonumber\\ &  -\left(\sum\limits_{\mathbf{O}\in\mathcal{O}_A}\#_{A}\mathbf{O}+\sum\limits_{\mathbf{K}\in\mathcal{K}_A } \indicator{\mathbf{K}< \mathbf{O}_1}\#_{A}\mathbf{K}\right)^2\\  
	&\	SED(A)=\frac{1}{n}\sum\limits_{i=1}^{n} \frac{sig(\#_{B_i}C_{out})\#_{B_i}F_{in}}{\#_{B_i}F_{out}}  \times pool\_SED(B_i)\times \\
	&~~~~~~~~~~~~~~~~~~~~~~  sig \left( skip\_SED(B_i) \times conv\_SED(B_i) \right) 	
\end{align}
\end{definition}

The main idea behind the construction of SED is to rank architectures according to the suppression of local entropy decrease. According to Proposition~\ref{pro2}, Proposition~\ref{pro3} and Proposition~\ref{pro4}, we design skip\_SED, conv\_SED and pool\_SED to measure the effect of different operations on local entropy decrease.  
For the architecture search spaces that exclude skip connection and pooling operations, we set skip\_SED and pool\_SED to an idential constant to ensure a smooth calculation of SED.

\section{Experiments and Results} 
The SED proxy is evaluated from the following aspects: (i) the comparison among zero-cost proxies in terms of computation speed; (ii)  the comparison among zero-cost proxies in terms of architecture selection; (iii). the comparison among zero-cost proxies in terms of the correlation with network performance;  (iv) the performance of a complete SED-based NAS in DARTS space.

	\subsection{Experimental configurations}
\label{Experimental configurations}
\textbf{Hardware.}  The proxies are evaluated on a device with an Inter Core i7-12700K CPU,  GeForce RTX 3090 GPU and 64GB RAM.

\textbf{Benchmarks.} We evaluate the proxies on five popular benchmarks in NAS studies as follows:
(i) \emph{NATS-Bench-TSS} \cite{RN416} contains 15,625 networks with different architectures. We compare SED to the other zero-shot proxies on  NATS-Bench-TSS with CIFAR-10 \cite{RN452}, CIFAR-100 \cite{RN452}, ImageNet16-120 \cite{RN456}. (ii) \emph{NATS-Bench-SSS} \cite{RN418} consists of 32, 768  architectures. We evaluate proxies on  NATS-Bench-SSS with CIFAR-10, CIFAR-100, and ImageNet16-120. (iii) and (iv) \emph{NAS-Bench-301} \cite{RN417}  and  \emph{EvoXbench-DARTS}\cite{RN424} are surrogate NAS benchmarks for the DARTS\cite{RN427} search space using CIFAR-10. (v) \emph{NAS-Bench-101} \cite{RN419} consists of 423,624 architectures evaluated on  CIFAR-10.  

\textbf{Baselines.}  Current SOTA and some competitive proxies are compared as baselines, including  meco \cite{RN411}, zico \cite{RN420}, zen \cite{ming_zennas_iccv2021}, grasp \cite{RN435}, synflow \cite{RN434} and grad\_norm \cite{RN432}.

\begin{table}
	\centering
	\caption{Computing hours required to evaluate NATS-Bench-TSS and NATS-Bench-SSS using CIFAR-10 and CIFAR-100}
	\label{Time_consume}
	\resizebox{\columnwidth}{!}{%
		\begin{tabular}{lcccc}
			\hline
			\multirow{2}{*}{Proxy} & \multicolumn{2}{c}{NATS-Bench-TSS} & \multicolumn{2}{c}{NATS-Bench-SSS} \\
			& CIFAR-10 & CIFAR-100 & CIFAR-10 & CIFAR-100 \\
			\hline
			meco & 3.29 & 3.51 & 2.29 & 1.92 \\
			zico & 2.47 & 2.69 & 1.43 & 1.32 \\
			synflow & 2.76 & 2.57 & 1.22 & 1.32 \\
			zen & 2.53 & 2.59 & 1.44 & 1.37 \\
			grad\_norm & 2.37 & 2.39 & 1.23 & 1.27 \\
			grasp & 3.01 & 3.34 & 2.18 & 1.61 \\
			\hline
			SED & \textbf{1.28e-4} & \textbf{1.28e-4} & \textbf{1.03e-4} & \textbf{1.03e-4} \\
			\hline
		\end{tabular}%
	}
\end{table}

\begin{table*}
	\centering
	\caption{ Optimal test accuracy and ranking of architectures with greatest proxy value on NATS-Bench-TSS and NATS-Bench-SSS using CIFAR-10, CIFAR-100, and ImageNet16-120, respectively.}
	\label{tab:combined_ranking_results}
	\resizebox{2\columnwidth}{!}{%
	\begin{tabular}{lcccccccccccc} 
		\hline
		\multirow{3}{*}{Proxy}                   & \multicolumn{6}{c}{NATS-Bench-TSS}                                                                                                                                                                                                                         & \multicolumn{6}{c}{NATS-Bench-SSS}                                                                                                                                                                                                \\ 
		\cline{2-13}
		& \multicolumn{2}{c}{CIFAR-10}                                                      & \multicolumn{2}{c}{CIFAR-100}                                                     & \multicolumn{2}{c}{ImageNet16-120}                                                 & \multicolumn{2}{c}{CIFAR-10}                                                      & \multicolumn{2}{c}{CIFAR-100}                                                     & \multicolumn{2}{c}{ImageNet16-120}                        \\ 
		\cline{2-13}
		& Test(\%)                                  & Rank                                  & Test(\%)                                  & Rank                                  & Test(\%)                                  & Rank                                   & Test(\%)                                  & Rank                                  & Test(\%)                                  & Rank                                  & Test(\%)         & Rank                                   \\ 
		\cline{2-13}
		\vcell{grasp}                            & \vcell{82.04}                             & \vcell{13659}                         & \vcell{60.01}                             & \vcell{11554}                         & \vcell{35.29}                             & \vcell{8510}                           & \vcell{86.11}                             & \vcell{17,572}                        & \vcell{45.43}                             & \vcell{26,388}                        & \vcell{25.29}    & \vcell{26980}                          \\[-\rowheight]
		\printcellbottom                         & \printcellmiddle                          & \printcellmiddle                      & \printcellmiddle                          & \printcellmiddle                      & \printcellmiddle                          & \printcellmiddle                       & \printcellmiddle                          & \printcellmiddle                      & \printcellmiddle                          & \printcellmiddle                      & \printcellmiddle & \printcellmiddle                       \\
		meco                                     & 93.76                                     & 166                                   & 71.11                                     & 173                                   & 41.44                                     & 2795                                   & 76.25                                     & 32,752                                & 31.07                                     & 32,607                                & 22.79            & 29928                                  \\
		zico                                     & 93.76                                     & 166                                   & 71.11                                     & 173                                   & 41.44                                     & 2795                                   & 88.67                                     & 353                                   & 60.33                                     & 17                                    & 37.36            & 64                                     \\
		synflow                                  & 93.76                                     & 166                                   & 71.11                                     & 173                                   & 41.44                                     & 2795                                   & 89.45                                     & 1                                     & 60.94                                     & 2                                     & 39.46            & 1                                      \\
		zen                                      & 90.7                                      & 7816                                  & 68.25                                     & 3290                                  & 40.60                                     & 3511                                   & 88.8                                      & 202                                   & 60.38                                     & 13                                    & 37.46            & 60                                     \\
		grad\_norm                               & 89.27                                     & 9691                                  & 47.61                                     & 14072                                 & 35.82                                     & 8208                                   & 86.69                                     & 12589                                 & 33.85                                     & 32,195                                & 18.33            & 32424                                  \\
		\#Param                                  & 93.76                                     & 166                                   & 71.11                                     & 173                                   & 41.44                                     & 2795                                   & 89.45                                     & 1                                     & 60.94                                     & 2                                     & 39.46            & 1                                      \\ 
		\hline
		\multicolumn{1}{c}{\textbf{SED (Ours) }} & \textbf{\textbf{\textbf{\textbf{94.36}}}} & \textbf{\textbf{\textbf{\textbf{3}}}} & \textbf{\textbf{\textbf{\textbf{73.51}}}} & \textbf{\textbf{\textbf{\textbf{1}}}} & \textbf{\textbf{\textbf{\textbf{46.34}}}} & \textbf{\textbf{\textbf{\textbf{24}}}} & \textbf{\textbf{\textbf{\textbf{89.45}}}} & \textbf{\textbf{\textbf{\textbf{1}}}} & \textbf{\textbf{\textbf{\textbf{60.94}}}} & \textbf{\textbf{\textbf{\textbf{2}}}} & \textbf{39.46}   & \textbf{\textbf{\textbf{\textbf{1}}}}  \\
		\hline
	\end{tabular}
}
\end{table*}
\subsubsection{Performance of SED on Computation Speed}
\label{Performance of SED on Computation Speed} 
The time required to evaluate the full architectures is summarised in Table~\ref{Time_consume} in NATS-Bench-TSS and  NATS-Bench-SSS using CIFAR-10 and CIFAR-100.  Experimental results show that SED computes   10,000$\times$ faster than the rest of zero-cost proxies for architectural evaluation.  Intuitively, SED completes the evaluation of architecture space with five nodes in Example~\ref{exam:1} in 0.1 hours.   In addition, the calculation of SED is performed on a single core of the CPU, resulting in  a six-order magnitude reduction in floating-point operations compared to other proxies.  Topology-based computing makes SED data-free and network-running-free, accelerating the computation.

	\subsubsection{Performance of SED on Architecture Selection}
\label{Performance of SED on Architecture Selection}
In terms of the performance of the selected architectures, we evaluate SED  on architecture topology spaces  (NATS-Bench-TSS, NAS-Bench-101, NAS-Bench-301 and EvoXBench-DARTS) and architecture size space (NATS-Bench-SSS). The performance of the architecture corresponding to the greatest proxy is presented in Table~\ref{tab:combined_ranking_results}. The average test accuracy of Top-10 architectures of these benchmarks is presented  \textbf{ in the Appendix (in Table~1)}, where SED achieves the highest average test accuracy of Top-10 architectures among the five benchmarks.  In terms of the average accuracy of Top-10 architectures, SED improves the average test accuracy by 0.51\% and 1.07\% on  DARTS space and NATS-Bench-TSS, respectively.  

According to Table~\ref{tab:combined_ranking_results},   SED exhibits competitive performance on these benchmarks in terms of the test accuracy of architecture with the greatest proxy value.  For example, by solely maximizing SED, the architectures ranked \textbf{1-st, 2-nd, and 1-st} are selected on NATS-Bench-SSS using CIFAR-10, CIFAR-100 and ImageNet16-120, respectively (\textbf{3-rd, 1-st, and 24-th }for NATS-Bench-TSS).  However, most of the other proxies select the suboptimal architectures by solely maximizing the proxy value because the corresponding architectural performance shows a phenomenon of initially improving and afterward reducing as the proxy value increases.    One of the reasons leading to this phenomenon is the similarity of the working mechanism between the existing proxies and \#Param, due to the strong correlation between proxy value and \#Param in the subsequent correlation comparison.  These experimental results show that SED is competent for architecture selection tasks in multiple architecture spaces.

	\subsubsection{Performance of SED on Spearman's $\rho$}
\label{Performance of SED on Spearman's}

The correlation between proxy value and the performance of architectures is evaluated on multiple benchmarks. The experimental results are presented  in the Appendix  (in Table~2 to Table~4),  where SED achieves the highest correlation on 4 out of 5 benchmarks. For instance, SED achieves Spearman's $\rho$ of 0.61 in the DARTS space, which is 0.09 higher than the SOTA zero-cost proxies. (\textbf{An over all heat map will be provided in the next version}).
 
We further evaluate Spearman's $\rho$ between minimum test accuracy of architecture with the same proxy value and SED on NATS-Bench-TSS. The results are summarized in  the Appendix. SED achieves Spearman's $\rho$ of 0.89, 0.85, and 0.88 on CIFAR-10, CIFAR-100, and ImageNet16-160, respectively. It means the minimum test accuracy of the selected architecture is highly correlated with SED. Moreover,  we compare the architecture performance with the same SED value. The results are summarized  in the appendix. According to the data,  architectures with the same SED value has similar performance (much smaller variance than randomly selected architectures). Therefore, the average, minimum and maximum test accuracy increase as the SED value increases, leading to select high-performance architectures, even with relatively low correlation.
 
The results in Table~\ref{tab:combined_ranking_results}  demonstrate that meco, zico, and synflow select the same architectures as \#Param does in NATS-Bench-TSS, which indicates that these proxies may work in a similar mechanism. 
Current SOTA proxies show high correlations with \#Param, which is consistent with the results in Table~\ref{tab:combined_ranking_results}. It indicates that the effectiveness of SOTA proxies may depend on the number of parameters. Meanwhile, SED shows a lower correlation with \#Param but selects the architectures with higher performance. The lower correlation indicates that SED works in a different mechanism from current proxies, which is complementary to current proxies.

\begin{figure}[!htp]
 	\centering
 	\includegraphics[width=0.81\columnwidth]{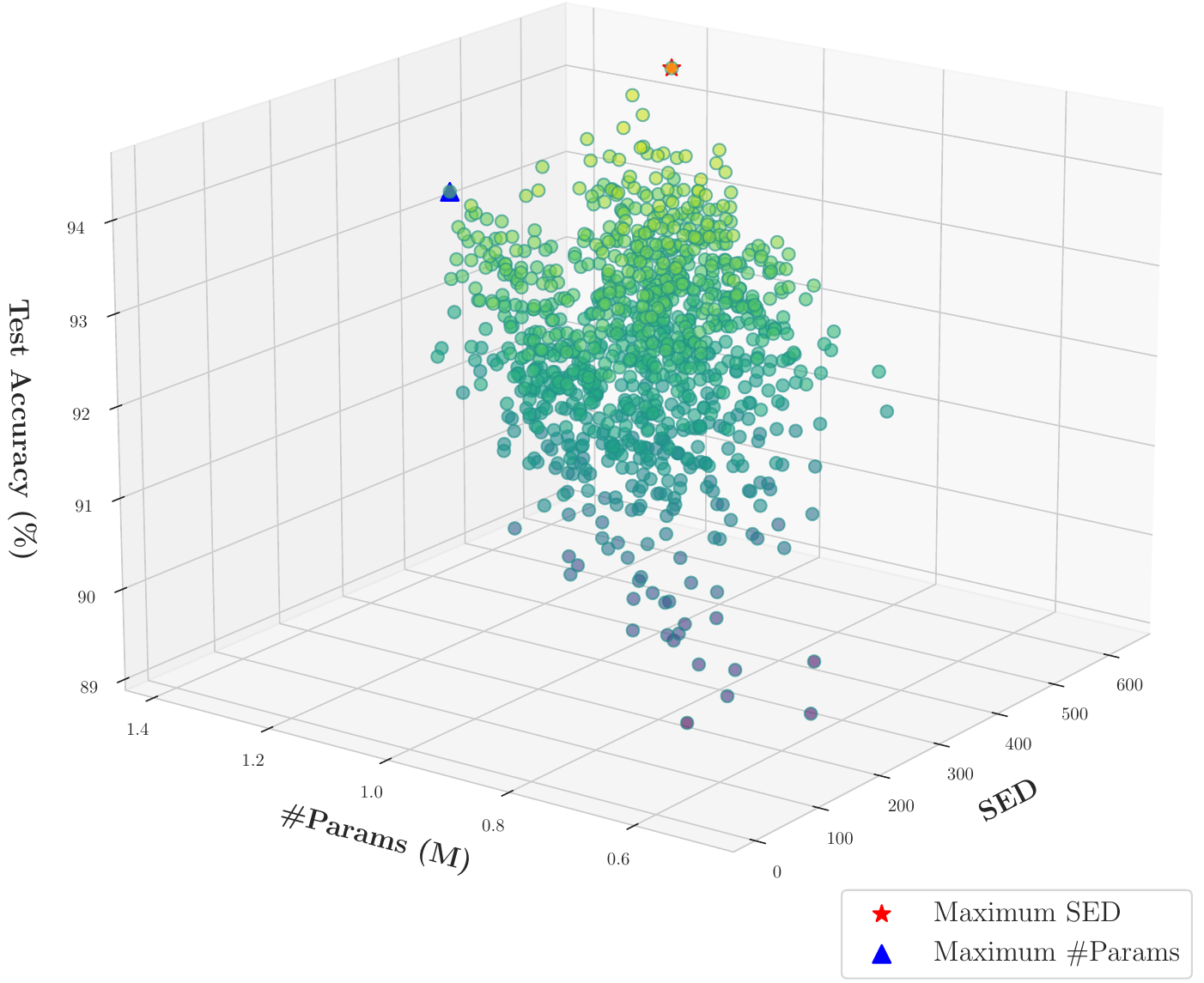}
 	\caption{Comparison between \#Param and SED  on DARTS space out of 1,000 randomized architectures.}
 	\label{fig:ParamVSSED}
 \end{figure}

  To further exhibit the difference between SED and \#Param,  we compare the relationship between SED-\#Param pairs and network performance. The results are visualized in Figure~\ref{fig:ParamVSSED}, in which the architectures are divided into two clusters. The separated data clusters provide visual evidence of the lower correlation between SED and \#Param.   As SED increases, \#Param first increases and then decreases, demonstrating that SED works in a different mechanism from \#Param. By solely maximizing proxy values, SED selects the architectures with 1.09\% higher test accuracy and 0.17M smaller than \#Params. The selected architectures are marked as red star and blue triangle in  Figure~\ref{fig:ParamVSSED}, respectively. The results in Figure~\ref{fig:ParamVSSED} show that the SED proxy works in a relatively new mechanism.   Meanwhile, under the working mechanism of SED the optimal architectures are selected. It demonstrates that the SED proxy is effective and complementary to current SOTA proxies.

\subsection{Performance of SED-based NAS}
\label{The performance of SED-based NAS}
 The SED proxy is utilized on randomized architecture to construct a complete NAS method. The architecture with the greatest SED value out of 2,000 randomized architectures is selected to train a network. Zero-Cost-PT \cite{Xiang2021ZeroCostPM} is used for other zero-cost proxies as a generation strategy to efficiently select architectures. DARTS \cite{RN427} is used as the search space to demonstrate the performance of proxies. The experimental settings are presented in the Appendix (in Table~7 and Table~8).

 We compare our SED-based method with current zero-shot NAS methods in the DARTS search space. The Experimental results in Table~\ref{tab:darts_result} demonstrate that the SED-based method prompts the baselines specifically; in terms of search cost, the SED-based method completes architecture selection in 0.3 CPU seconds. The search cost of the SED-based method is negligible compared to the compared methods.

 In terms of the performance of selected architectures, the experimental results on CIFAR-10, CIFAR-100 and ImageNet1K \cite{RN579} are summarized in Table~\ref{tab:darts_result}. The SED-based method outperforms the baselines of zero-shot methods that use a single proxy. With 0.3M parameters less, the architecture selected by SED-based NAS reduces the Top-1 error by 0.03\% on CIFAR-10. Meanwhile, the Top-1 accuracy is improved by 0.42\% and 0.61\% on CIFAR-100 and ImageNet1K, respectively. The selected architectures are presented  in the Appendix (in Figure~3 and Figure~4).   
 These experimental results on complete NAS verify the effectiveness of SED. Especially, SED consumes much less time to select a better architecture than baselines.  
\textbf{ (The experimental results on transformer space will be supplied in the next version)}

\subsection{More Experimental Results in the Appendix}
As for more experiments, the effectiveness of the local one-dimensional entropy is verified, and the results of 35 out of 36 sets of experiments satisfy the theoretical predictions of the proposed propositions. Meanwhile, we conducted ablation experiments, whose results indicate that the suppression of local entropy causes the reduction of network performance. The detailed experimental results are presented in the Appendix ( in Table~5 and Figure~2).

\begin{table}
	\centering
	\caption{ Comparisons of SED-based NAS with baselines on DARTS}
	\label{tab:darts_result}
	\resizebox{\columnwidth}{!}{%
	\begin{tabular}{llll} 
		\hline
		\multicolumn{1}{c}{Approach}     & \multicolumn{1}{c}{\begin{tabular}[c]{@{}c@{}}Test Error \\(\%)\end{tabular}} & \multicolumn{1}{c}{\begin{tabular}[c]{@{}c@{}}Search Cost \\(GPU hours)\end{tabular}} & \multicolumn{1}{c}{\begin{tabular}[c]{@{}c@{}}Params \\(M)\end{tabular}}  \\ 
		\hline
		\multicolumn{4}{c}{\textbf{CIFAR-10 Results}}                                                                                                                                                                                                                                        \\ 
		\hline
		Zero-Cost-PT$_\text{synflow}$    & 2.96 $\pm$ 0.11                                                               & 0.72                                                                                  & 5.1                                                                       \\
		Zero-Cost-PT$_\text{zen}$        & 2.85~$\pm$ 0.01                                                               & 1.2                                                                                   & 3.9                                                                       \\
		Zero-Cost-PT$_\text{grasp}$      & 2.73 $\pm$ 0.10                                                               & 2.4                                                                                   & 3.3                                                                       \\
		Zero-Cost-PT$_\text{grad\_norm}$ & 2.93 $\pm$ 0.17                                                               & 0.96                                                                                  & 3.4                                                                       \\
		Zero-Cost-PT$_\text{zico}$       & 2.80 $\pm$ 0.03                                                               & 0.96                                                                                  & 5.1                                                                       \\
		Zero-Cost-PT$_\text{meco}$       & 2.69$\pm$ 0.05                                                                & 1.92                                                                                  & 4.2                                                                       \\ 
		\hline
		SED (\textbf{Ours})              & \textbf{2.66} $\pm$ \textbf{0.05}                                             & \textbf{0.00028}                                                                      & \textbf{3.9}                                                              \\ 
		\hline
		\multicolumn{4}{c}{\textbf{CIFAR-100 Results}}                                                                                                                                                                                                                                       \\ 
		\hline
		Zero-Cost-PT$_\text{synflow}$    & 19.82$\pm$ 0.35                                                               & 0.96                                                                                  & 1.2                                                                       \\
		Zero-Cost-PT$_\text{zen}$        & 21.35$\pm$ 0.29                                                               & 1.38                                                                                  & 0.9                                                                       \\
		Zero-Cost-PT$_\text{grasp}$      & 22.65$\pm$ 0.30                                                               & 3.12                                                                                  & 0.7                                                                       \\
		Zero-Cost-PT$_\text{grad\_norm}$ & 23.11$\pm$ 0.35                                                               & 0.96                                                                                  & 0.8                                                                       \\
		Zero-Cost-PT$_\text{zico}$       & 19.54$\pm$ 0.28                                                               & 1.44                                                                                  & 1.1                                                                       \\
		Zero-Cost-PT$_\text{meco}$       & 19.33$\pm$ 0.25                                                               & 1.92                                                                                  & 0.8                                                                       \\ 
		\hline
		SED (Ours)                       & \textbf{18.91} $\pm$ \textbf{0.22}                                            & \textbf{0.0002}                                                                       & \textbf{1.0}                                                              \\ 
		\hline
		\multicolumn{4}{c}{\textbf{ImageNet1K (searched on CIFAR-100)}}                                                                                                                                                                                                                      \\ 
		\hline
		Zero-Cost-PT$_\text{zico}$       & \multicolumn{1}{c}{26.51}                                                     & \multicolumn{1}{c}{1.44}                                                              & \multicolumn{1}{c}{6.7}                                                   \\
		Zero-Cost-PT$_\text{meco}$       & \multicolumn{1}{c}{26.69}                                                     & \multicolumn{1}{c}{1.92}                                                              & \multicolumn{1}{c}{4.8}                                                   \\ 
		\hline
		SED (Ours)                       & \multicolumn{1}{c}{\textbf{25.90 }}                                           & \multicolumn{1}{c}{\textbf{0.0002}}                                                   & \multicolumn{1}{c}{\textbf{6.0 }}                                         \\
		\hline
	\end{tabular}
}
\end{table}

\section{Conclusion}
In this study,  we demonstrate that irrational architectures will decrease local entropy. Meanwhile, the reduction in local entropy degrades network performance. Therefore, the SED proxy is proposed to select the architectures that highly suppress local entropy decrease. Compared with previous work, SED computations rely only on architecture topologies, making SED thousands of times faster than existing proxies. Experimental results on multiple public benchmarks show that SED significantly outperforms the SOTA zero-cost proxies in terms of architecture selection and computation speed. Theoretical and experimental results in this study verify that architecture topologies are highly correlated with network performance.

\bibliography{Reference.bib}

\end{document}